# A Holistic Assessment of the Carbon Footprint of Noor, a Very Large Arabic Language Model

**Imad Lakim**
TII, Abu Dhabi
imad.lakim@tii.ae

**Ebtesam Almazrouei**
TII, Abu Dhabi
ebtesam.almazrouei@tii.ae

**Ibrahim Abu Alhaol**
TII, Abu Dhabi
ibrahim.abualhaol@tii.ae

**Merouane Debbah**
TII, Abu Dhabi
merouane.debbah@tii.ae

**Julien Launay**
LightOn, Paris
julien@lighton.io

## Abstract

As ever larger language models grow more ubiquitous, it is crucial to consider their environmental impact. Characterised by extreme size and resource use, recent generations of models have been criticised for their voracious appetite for compute, and thus significant carbon footprint. Although reporting of carbon impact has grown more common in machine learning papers, this reporting is usually limited to compute resources used strictly for training. In this work, we propose a holistic assessment of the footprint of an extreme-scale language model, Noor. Noor is an ongoing project aiming to develop the largest multi-task Arabic language models–with up to 13B parameters–leveraging zero-shot generalisation to enable a wide range of downstream tasks via natural language instructions. We assess the total carbon bill of the entire project: starting with data collection and storage costs, including research and development budgets, pretraining costs, future serving estimates, and other exogenous costs necessary for this international cooperation. Notably, we find that inference costs and exogenous factors can have a significant impact on total budget. Finally, we discuss pathways to reduce the carbon footprint of extreme-scale models.

## 1 Introduction

Recent progress in natural language processing (NLP) has been driven by the emergence of so-called foundation models (Bommasani et al., 2021). This paradigm shift is characterised by a homogenisation of modelling methods– crystallising around the Transformer architecture (Vaswani et al., 2017)– and by emergent capabilities (e.g. zero-shot generalisation) predominantly arising from sheer scale alone (Brown et al., 2020). NLP models are now experiencing a 3-4 months doubling time in size, as outlined by Figure 1. Most recent large language models such as MT-NLG 530B (Smith et al., 2022), Gopher 280B (Rae et al., 2021), or Jurassic-1 178B (Lieber et al., 2021), all report training budgets in the thousands of PF-days[1] range. Because AI accelerators performance per watt has plateaued compared to deep learning budgets (Reuther et al., 2021; Sevilla et al., 2022), practitioners have had to scale-out training over an increasingly large number of accelerators (Narayanan et al., 2021). Accordingly, the energy cost of training state-of-the-art models has grown significantly: increase in compute is no longer fuelled by improvements in hardware efficiency, but in hardware scale.

Although this increase in size and compute budget is backed by empirical scaling laws drawing a clear link between compute spent and model performance (Kaplan et al., 2020), the societal benefits of larger models have been questioned (Tomašev et al., 2020; Bender et al., 2021). Specifically to environmental concerns, in a time of climate crisis when carbon emissions must be drastically cut (Masson-Delmotte et al., 2018), one may question whether these large compute budgets are justified. A crucial step towards answering this question is an in-depth evaluation of the footprint of large models.

Existing assessments of the environmental impacts of large models are usually focused on hyperparameter tuning and pretraining costs (Strubell et al., 2019; Patterson et al., 2021). This trend is reflected by the growing number of tools available to help practitioners quantify the impact of machine learning computations (Bannour et al., 2021). If some studies have also endeavoured to quantify select aspects of the machine learning pipeline (e.g. conference attendance (Skiles et al., 2021), hardware lifecycle (Gupta et al., 2021), etc.), end-to-end evaluations of machine learning projects life cycle emissions remain rare (Wu et al., 2022).

[1] A PF-day is 1 PFLOPs (10 A100) sustained for a day.

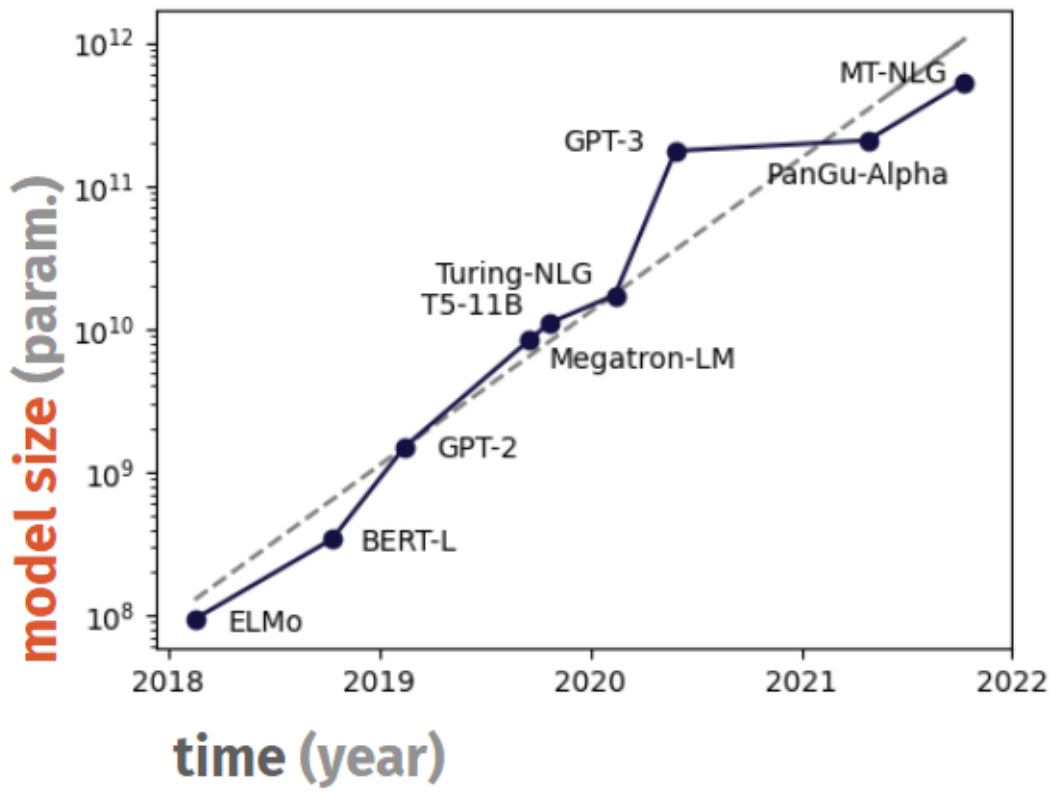

Figure 1: **Over the last four years, the size of state-of-the-art language models has doubled every 3-4 months.** Note that this trend has been slowing down, due to scale-out limitations.

To fill this gap, we produce an end-to-end assessment of the carbon footprint of Noor, a project seeking to train a very large Arabic language model. Our contributions are the following:

**Holistic assessment.** We evaluate the total carbon bill of the entire project: starting with data collection, curation, and storage, including research and development and hyper-parameters tuning budgets, pretraining costs, future serving estimates, and other exogenous impacts sparked by this international cooperation (e.g. flights, personnel, etc.)

**Beyond pretraining.** We identify pretraining compute as driving more than half of the emissions of the project. However, all combined, other R&D, storage, and personnel counts still amount for 35% of the carbon footprint. We also identify downstream use in the wild as potentially significant. This leads us to recommend for the end-to-end footprint to be systematically assessed on a per-project basis. Notably, in scenarios with a low-impact training electric mix, costs beyond pretraining may become the main sources of emissions.

**Pathways to lower footprints.** Finally, we discuss ways to reduce the environmental footprints of projects involving large models, and put in perspective the footprint of similar projects.

## 2 Related work

In light of ever increasing computational budgets (Sevilla et al., 2022) and of the need to cut on emissions to abate global warming (Masson-Delmotte et al., 2018), the environmental impact of deep learning has drawn significant interest.

Strubell et al., 2019 notably highlighted the potential high environmental costs of deep learning. However, its headline figures were produced in the specific context of neural architecture search, a relatively rare practice for extreme-scale models nowadays. Lacoste et al., 2019; Lottick et al., 2019; Schwartz et al., 2020 subsequently called for AI research to be more aware of its environmental cost. An increasing number of tools, such as `codecarbon` (Schmidt et al., 2021), have been developed to help with tracking the impact of deep learning experiments (Bannour et al., 2021). All of these lines of research share similar recommendations: the carbon footprint of deep learning is a direct consequence of the electricity mix and efficiency of the data center, suggesting that picking an appropriate provider is the most straightforward way to reduce environmental impact.

Specifically to extreme-scale models, Patterson et al., 2021 estimated the energy consumption of five large NLP models, including GPT-3. They identified that a judicious choice of neural architecture, datacenter and accelerator can help reduce considerably carbon budgets. Thompson et al., 2020 identified a clear relationship between large models performance and their carbon impact, building upon work on neural scaling laws (Kaplan et al., 2020). Taddeo et al., 2021 estimated the cost of training GPT-3 in different data centers across the worldwide, highlighting again the high dependency on the local energy mix and specific infrastructure.

Two recent studies have provided insights into the end-to-end carbon footprint of deployed models in the industry. Wu et al., 2022 studied the impact of the increasingly large recommender systems leveraged at Meta, while Patterson et al., 2022 provided an assessment of the costs (including inference) of large models at Google. They expect the carbon footprint of training to plateau in coming years, and then to shrink–owing to more efficient high performance computing platforms. They also assert that current studies are overestimating the real environmental costs of large models, in light of the wide availability of ”clean” compute platforms.

In the field of astrophysics, Aujoux et al., 2021 did an extensive study to estimate the carbon footprint of the Giant Array for Neutrino Detection (GRAND) project, a multi-decade worldwide experiment. Inspired by their holistic methodology, we seek to establish the first end-to-end assessment of an extreme-scale NLP project.

## 3 The Noor Project

The current state-of-the-art generative language model in Modern Standard Arabic is AraGPT (Antoun et al., 2021), a 1.5B parameters model. The Noor project seeks to expand upon this model, introducing a 1.5B, 2.7B, 6.7B, and 13B Arabic models, trained a custom curated dataset of 150B tokens, inspired by The Pile (Gao et al., 2020). These larger scales are expected to make the model able to tackle novel tasks through zero-shot generalization, as exhibited by GPT-3 (Brown et al., 2020) or GPT-J (Wang and Komatsuzaki, 2021).

Noor is an on-going international cooperation between the Technology Innovation Institute in the United Arab Emirates and LightOn in France. The Noor project can be split in four parts:

- **Data curation.** A custom curated dataset of 150B tokens has been assembled for Noor. This dataset has been scrapped from diversified sources, and also includes data from Common Crawl. We filter this data with an LM-based quality-scoring system inspired by CCNet (Wenzek et al., 2019).
- **R&D experiments.** To validate tokenization, dataset, architecture, and establish scaling laws, we trained a number of R&D models (100M-1.5B parameters on 10-30B tokens).
- **Main training.** We train a suite of four models of 1.5B, 2.7B, 6.7B, and 13B parameters.
- **Model use.** Prospectively, we include some estimations of the future inference cost of these models as they are put in use.

## 4 Factors influencing the carbon footprint of large models

Before beginning our assessment, we propose to identify some of the key influencing factors on the potential carbon footprint of large models, focusing first on factors directly related to the models themselves and not to the project producing them.

**Model size.** The number of floating operations per forward pass is directly proportional to the size of the network. A common approximation for the total compute budget $C$ required for training a Transformer model with $N$ parameters on $D$ tokens is $C = 6ND$ (Kaplan et al., 2020). As the optimal dataset size only grows sublinearly with model size for autoregressive modelling (Henighan et al., 2020), compute budget will scale more or less linearly with model size. The larger the number of operations, the more energy is needed to train the model. For inference, the cost for each token is reduced to a third compared to training, and environmental impact will be driven by the total number of words/tokens processed.

**Hardware characteristics.** The throughput (in FLOPs) that can be tackled by the hardware will drive the total time required to perform the task. More efficient hardware will have more throughput per Watt. We note however that most available chips suitable for large model training (e.g., NVIDIA GPUs, Google TPUs, etc.) exhibit similar efficiency characteristics (Reuther et al., 2021).

**Modelling decisions.** We identified above two key factors: number of tokens processed (for training or inference), and hardware throughput. We note that both of these are also strongly impacted by modelling decisions. A more fertile tokenizer will use less tokens for the same text, leading to faster processing. Similarly, small changes in model architecture (e.g., choosing hidden sizes in accordance with wave/tile quantization) and in implementation (e.g., 3D parallelism) can drastically increase throughput, and reduce total training time.

**Data center efficiency.** The energy consumed does not serve only to power up the servers, but also to cool down the data center itself and to respond to other electrical needs. The Power Usage Effectiveness (PUE) is used to assess the overall efficiency of a data center. It measures the quotient of the total energy requirement and the final energy used by the servers. The PUE will be influenced by the data center architecture. Worldwide average is around 1.8, but Google for instance reports an average PUE of 1.11. Waste heat in data centers can also be reused for collective water heating, driving down the PUE, as in the Jean Zay HPC.

**Electricity mix.** The breakdown of the energy sources powering a data center is a crucial factor, and depends primarily on the region. The electricity mix determines the carbon emissions per kWh of electricity. Today, the world average of carbon emission by kwh of electricity generated is 475 gCO2e/kWh, and an increasing number of data centers from cloud providers are using 100% renewable or nuclear energy to power their hardware. Taking Google Cloud as an example again, their Montreal facility reports 27gCO2e/kWH, twenty times lower than the world average.

Beyond factors related to the models themselves, we seek in this study to take into account a number of other costs: storage, preprocessing, and transfer costs for the dataset, personnel costs such as travel and individual laptops, etc. We note however one limitation from our study: we do not take into account the lifecycle of the hardware used. Unfortunately, numbers are scarcely available, and not made public by the main manufacturers.

## 5 Carbon footprint of the Noor project

### 5.1 Electricity consumption

We begin by accounting for the electricity consumption of all aspects of the project. The impact of this consumption will be highly dependent on the carbon intensity of the electricity mix used. Non-electric sources (e.g., international flights) will be added to the carbon budget in a second phase.

#### 5.1.1 Data storage and transfers

The energy consumption of data depends on both the energy required for powering the disks to store the data, and the energy consumed when moving the data from one server to another. We average storage costs over the 6 months of the project.

**Storage.** Although disk wattage is generally reported on per-disk level, Posani et al., 2019 estimates the power per TB of data using aggregated technical specifications. The paper reports that the average peak consumption of cloud storage is around 11.3W/TB. It means an energy consumption of 99 kWh/TB a year. This estimation considers a PUE of 1.6 and a redundancy factor of 2 since managed services will also have a back-up.

The breakdown of our data storage is as follows:

- **Curated data.** Including both raw and processed data, we have accumulated around 2TB of curated data. This is stored for the 6 months of the project, resulting in 99kWh used.

- **Bulk data.** We use Common Crawl (CC) for acquiring large amounts of web data. Each CC dump is on average around 10TB, and we discard it immediately after processing it. On average, it takes 24 hours to fully process a dump: we used 21 dumps from CC, meaning we stored 210TB of data for 24hours, equivalent to 57 kWh of energy consumption. After processing the dumps, we got on average 1.2TB of data per dump, thus 25TB in total. Considering that this data will be stored for 6 months, we end up with 1.3 MWh of energy consumption for the bulk data. Note that we keep the processed data in all languages (not just Modern Standard Arabic).

- **Models.** The weights of the Noor models (1.3B, 2.7B, 6.7B and 13B) are respectively 2.6GB, 5.4G, 13.4GB, and 26GB in half-precision. This corresponds to training checkpoints (including the full-precision optimizer) of 20.8GB, 43.2GB, 107.2GB, and 208GB. We save such checkpoints every 10B tokens. In total, we end-up with 5.7TB of model weights and intermediary checkpoints for future analysis and interpretability work, consuming 0.3MWh in total.

**Transfers.** Posani et al., 2019 provided an estimate of 23.9 kJ per GB (6.38 kWh per TB) transferred, using the formula of Baliga et al., 2011 and the same hypothesis as Aslan et al., 2017 (800km average distance between core nodes). The 210TB of CC data are downloaded on the preprocessing servers once; the 25TB of processed data are moved once to our archival machines, and another time to the HPC used for training; the curated data is downloaded once, moved to the archival machines, and then moved to the HPC; the 5.7TB of models are moved once from our HPC, and then to our inference servers for final models or to workstations for intermediary checkpoints. Consequently, we estimate the transfer energy bill at 1.8 MWh.

**Total.** Thus, the total energy consumption of data is estimated to be about <u>3.5 MWh</u>, dominated by the multilingual Common Crawl data. We note that as ideal dataset size increases sublinearly with model size (Kaplan et al., 2020), we expect checkpoints and model transfers to eventually dominate the costs of storage and transfer for larger models.

Note that we neglect costs linked to a potential public release of the models, as it is difficult to predict traffic. As a rough estimation, 10,000 downloads of the 13B model would represent 260TB of traffic, and 1.66MWh consumed.

#### 5.1.2 Data processing

We take all text data through a pipeline inspired by CCNet (Wenzek et al., 2019) for preprocessing. This pipeline takes care of deduplication, language identification, and finally quality filtering with a reference language model trained on Wikipedia. Processing with our pipeline occurs on a CPU cluster with 768 cores, split over 16 nodes.

Table 1: **Training compute budget and energy used for training the Noor models.** Assuming a pretraining dataset of 150B tokens and a throughput of 100 TFLOPs per A100.

| **Model** | **Budget** [PF-days] | **Budget** [A100-hours] | **HPC** | **Consumption** [MWh] |
|---|---|---|---|---|
| 1.3B | 13.5 | 3300 | MeluXina | 2.1 |
| 2.7B | 28.1 | 6800 | Noor-HPC | 4.8 |
| 6.7B | 69.8 | 17000 | Noor-HPC | 11.8 |
| 13B | 135 | 33000 | Noor-HPC | 22.9 |

Using average high-performance CPUs TDP figures, we estimate the average power consumption of each node at 350W; hence, the power of the cluster is 5.6kW. We processed 21 dumps of CommonCrawl, plus our curated data, for a total of 381 wall-clock hours. Accordingly, assuming a PUE of 1.1 as reported by Google, the total energy consumed by data preprocessing is 2.35MWh.

Note that for CommonCrawl data, this results in data processed for every language supported (176 for identification, 48 for quality filtering). Accordingly, this cost could be amortised over future projects. For high-resource languages, this also results in very large amounts of data: processing more dumps would not be necessary, even to train a 1 trillion parameters model.

#### 5.1.3 Research and development

We carried experiments to validate tokenization methods, dataset composition, tune hyperparameters, and establish scaling laws. This early research and development work was performed on MeluXina, a high-performance super-computer located in Luxembourg. We used a total of 16,800 A100-hours in this phase. Each node used in MeluXina has 4 A100 SXM 40GB with a TDP of 400W, and two AMD EPYC 7763 CPUs with a TDP of 280W. They report a PUE of 1.35. Thus, we estimate the consumption of this R&D phase to be of 10.7MWh.

We expect the budget of this phase to roughly scale with model size. Indeed, debugging potential issues (e.g., numerical instabilities (Kim et al., 2021), etc.) for the final larger model will cost significantly more.

#### 5.1.4 Main training

Using the $C = 6ND$ approximation, it is possible to calculate in advance the training budget required for a specific model. We observe an effective throughput with our Megatron+DeepSpeed codebase of around 100 TFLOPs[2] across models, in line with the state-of-the-art. We train four main models (1.5B, 2.7B, 6.7B, 13B) on 150B tokens.

We train the smaller model on MeluXina, but the other three on our own HPC cluster. Each node contains 8 A100 80GB and 2 AMD EPYC 7763 CPUs. The PUE of our data center is 1.5, 20% more efficient than the world average.

Table 1 outlines the costs of the main training. The total electric energy consumed to train the Noor suite of models is thus 41.6 MWh, 55% of it spent on the largest 13B model.

#### 5.1.5 Inference

As the models of Noor have yet to be deployed, this is only a prospective estimate. Inference costs in general are difficult to estimate in advance, even more so for open source models which will be deployed to platforms with varying characteristics. We provide an estimate of the energy consumption during inference per generated token.

We thereafter denote as *processed tokens* the tokens in the original prompt sent to the model, and as *generated tokens* the tokens generated by the model using the prompt. To simplify calculations, we make the following assumptions from our experience with another large-scale API: (1) an A100 is used, which is sufficient for Noor-13B, but could be reduced to a more efficient T4 for Noor-1.5B/2.7B; (2) inference time per generated token is constant, whichever the number of processed tokens (per our benchmarks, thanks to caching, this is true up to 512 processed tokens roughly); (3) batch size is assumed to be 1, as batching is more challenging and less consistent for inference workloads.

Under these hypothesises, an A100 can generate up to 72,000 tokens per hour. Accordingly, we estimate that 26 Joules are required per token generated (400W for the GPU, 70W for the CPU, and 1.1 PUE on Google Cloud imply 517Wh of energy consumption for 72,000 tokens. Converted to Joule, it results in 26 Joules per token.) Accordingly, 3 billion tokens would have to be generated for inference costs to catch up with training costs. At some point during its beta, GPT-3 was reported to generate 4.5 billion words per day (Pilipiszyn, 2021).

[2] These are effective FLOPs for training the model, not hardware FLOPs. Hardware FLOPs are closer to 150 TFLOPs.

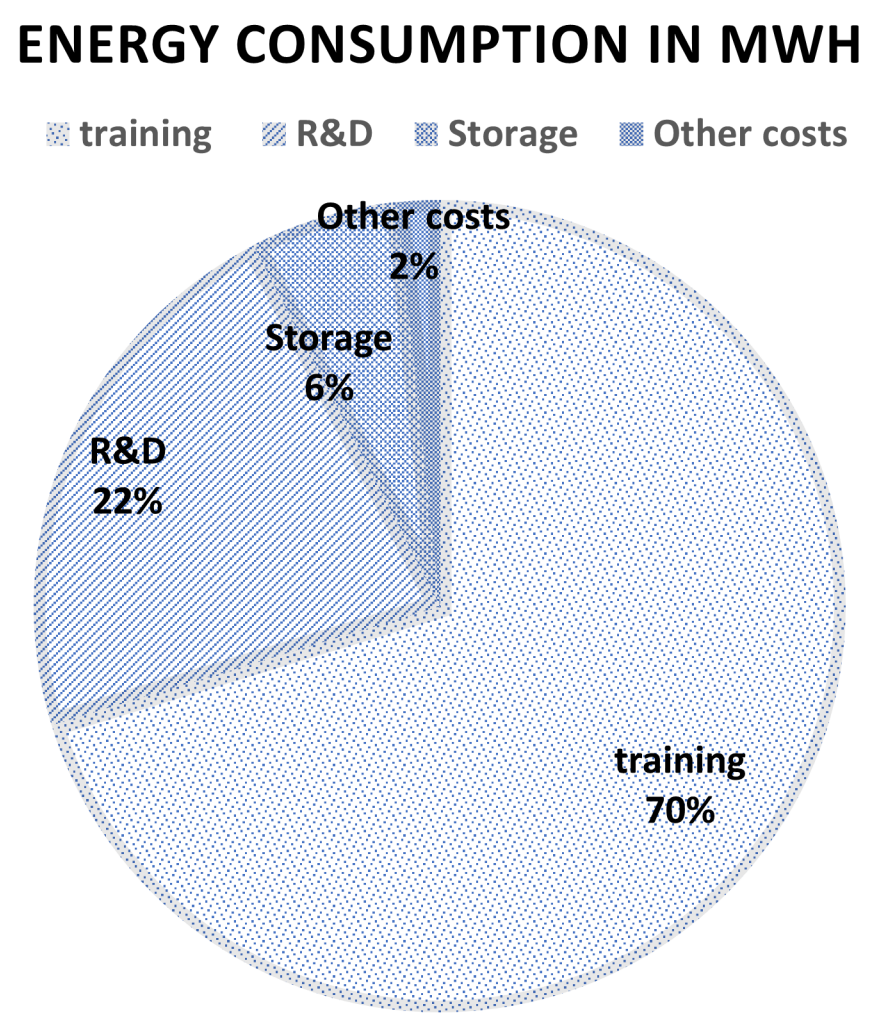

Figure 2: **Breakdown of the electricity consumption (total 59.14 MWh) of the Noor project.** Data preprocessing is included in R&D, amounting for 20% of it. We also note that R&D and dataset costs could be amortised through other projects or larger models.

#### 5.1.6 Additional costs

Beyond costs related to data, R&D, training, and inference, one may wonder if direct electricity use from scientists involved in the project is significant. Assuming that the average laptop consumes 70W, plus 30W for an external screen, six research scientists dedicating 100% of their time during 6 months for this project, 8 hours per day, will use up 0.604MWh. We could also include costs of e-mail exchanges and video-conferences specifically, but these were found to be negligible in Aujoux et al., 2021. We round up the marginal costs to 1MWh, and note that this is but a rough estimate.

#### 5.1.7 Summary

We showed that the total electricity consumption of the Noor project is not only about training the final models, as outlined in Figure 2. Nearly a third of the energy consumed (30%) went to tasks outside of main models pretraining.

Because of larger uncertainties, we keep the serving/inference assessment out of the previous budget. However, especially in the context of openly available models, the inference budget can rapidly catch up with the total budget outlined in 2.

### 5.2 Carbon footprint

Now, from the electricity consumption, and using information on the local carbon intensity, we will derive the full footprint of the Noor project. We will also add energy use coming from non-electric sources (e.g., flights). As the carbon intensity of the electricity mix varies significantly across regions, we outlined below the locations of interest:

- **Storage.** We used Amazon S3 in Bahrain;
- **R&D.** We used a GCP CPU cluster located in Netherlands, and MeluXina in Luxembourg;
- **Main training.** The smaller 1.3B model was trained on MeluXina, and the remaining models were trained on our dedicated HPC platform in the United Arab Emirates (UAE);
- **Other.** Six full-time scientists were involved, half in France and half in the UAE.

Table 2 shows the resulting carbon footprint for each of the development stages of Noor project. This highlights the importance of location for carbon footprint: notably, all calculations on performed on the relatively low-carbon MeluXina HPC end-up having very limited costs, even compared to small items like storage in Bahrain.

In addition to these development costs, we consider the carbon footprint of three round-trip flights of four scientists between Paris and Abu Dhabi. These trips were taken to run training workshops, brainstorming sessions, and discussions related to the project. We use the carbon emissions simulator of the International Civil Aviation Organization. One round-trip emits 527 kgCO2e per person, totalling 6.4 tons of emissions over all trips.

Finally, Figure 3 displays the total distribution of the carbon footprint of the project. As shown in the figure, factors like flights may be usually neglected, but have a significant contribution in the total carbon footprint. Specifically, as conference returns in-person, this is a systematic impact that exists on most papers. In the case of Noor, the few flights operated account for 18% of the total carbon emission of the whole project.

Table 2: **Carbon footprint of each phase of the Noor project.**

| Phase | Provider | Location | **Mix** [gCO2e/kWh] | **Use** MWh | **Footprint** [tCO2e] |
|---|---|---|---|---|---|
| **Storage** | Amazon S3 | Bahrain | 1188 | 3.5 | 4.2 |
| **R&D** | GCP | Netherlands | 410 | 2.35 | 0.96 |
| | MeluXina | Luxembourg | 60 | 10.7 | 0.65 |
| **Training** | MeluXina | Luxembourg | 60 | 2.1 | 0.13 |
| | Noor-HPC | UAE | 600 | 39.5 | 23.7 |
| **Others** | | France | 56 | 0.33 | 0.02 |
| | | UAE | 600 | 0.66 | 0.4 |

Interestingly, we note that with increasingly clean electricity and efficient data centers, the exogenous costs linked to flights and personnel are bound to increase in proportional impact.

**Inference.** Forecasting the carbon footprint of inference is harder for open models: as they may be downloaded and deployed by anyone, it is impossible to predict the carbon intensity of the electricity they will use. We study two scenarios: an intermediate one, based on the world average emission per kWh (475 gCO2e/kWh) and a best-case one, based on the low-impact French mix (56 gCO2e/kWh). These two scenarios correspond to around 300,000 tokens generated per kgCO2e, or to 2,500,000 tokens generated per kgCO2e in the best-case. Going back to the 4.5 billion words per day of GPT-3, this amounts to 30 tons of CO2e per day and 3.5 tons.

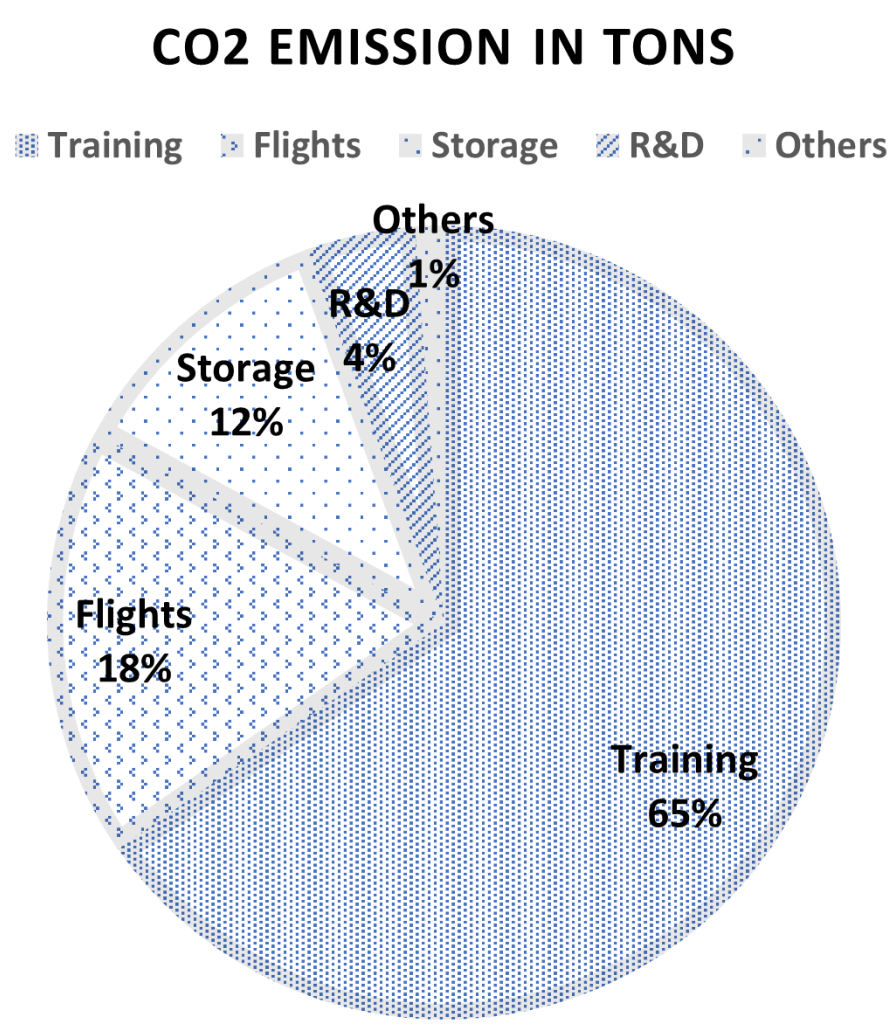

Figure 3: **Breakdown of the carbon footprint (total 36.5t tC02e) of the Noor project.** This breakdown is highly dependent on the localisation of the workloads and the local carbon intensity of the electricity mix.

## 6 Best practices and recommendations

From our experience with Noor, we highlight some recommendations for future projects to minimise their carbon footprint.

### 6.1 Modelling & engineering

A first angle of attack is to make the machine learning techniques used more efficient.

- **Efficient architectures.** Mixture-of-experts (MoE) models split the large fully-connected layers of a Transformer into distinct experts (Fedus et al., 2021). Although larger, MoE Transformers can bring significant energy savings during training and inference (Du et al., 2021), as the experts are only sparsely activated. Recent work demonstrate that they may even scale favorably compared to dense models (Clark et al., 2022). More broadly, even small changes (e.g. better embeddings, activation functions) may have a non-negligible impact on the overall carbon footprint.

- **Efficient inference.** As we have shown, inference costs can rapidly catch up with training costs: it is also interesting to make the model leaner for inference. Quantization (Yang et al., 2019) reduces numerical precision at inference time and accelerates inference, but it has seen limited adoption with large models. Distillation (i.e., training a smaller model from the outputs of a larger one) is a promising direction, already demonstrated for Transformers applied to vision (Touvron et al., 2021).

- **Efficient implementations.** Crucially, distributed training implementations must be as efficient as possible, to amortise the large idle consumption of the hardware – MeluXina reports for instance idle power of around 150W per GPU when accounting for CPU cores, infrastructure, etc. This includes taking into account fine-grained effects depending on architectures, such as wave and tile quantization, to achieve the best throughput possible.

### 6.2 Hardware

A second angle of attack is to focus on the hardware used to train these models.

- **Data center choice.** A data center with a PUE of 1.1 will decrease energy consumption by 39% compared to the world average of 1.8. Low PUE platforms should be preferred.

- **Local carbon intensity.** As highlighted by Table 2, the carbon intensity of the electricity mix significantly impacts the final footprint. Locating training in an area with a clean mix is an easy step to take that can drastically cut the footprint of a project. This is especially easy to do on online cloud platforms, which have many areas of availability.

- **Efficient inference.** Carefully selecting a proper AI accelerator for managed inference workloads can limit the footprint of model use. For instance, for smaller models (<3B), it may be possible to use T4s rather than A100s, which are 20% more energy efficient per FLOP than A100s. Finally, specialised accelerators are also starting to become available (Reuther et al., 2020). We note that this may however require specific developments.

### 6.3 Other practices

Finally, it is important to not underestimate costs beyond machine learning workloads.

- **Minimising exogenous impact.** Although we found the final footprint to be dominated by the main training runs, we still note the significant impact of the international flights taken during this cooperation (20% of the final footprint). Minimising such high-intensity cost center is important.

- **Costs reporting and offset.** The full cost of model development is rarely, if ever, reported in the literature. We highly recommend the AI community to start reporting the full energy consumption and the CO2e of their projects. This reporting can also be used as the basis for offsetting carbon emissions.

## 7 Discussion and conclusion

We undertook an end-to-end assessment of the carbon footprint associated with the development of an extreme-scale language model. We took into account data collection and storage, research and development, pretraining, and included estimates for future serving and inference. We also added personnel costs, such as international flights to run training workshops and brainstorming sessions.

In total, we estimate the development of the suite of the four Noor models to have emitted 36.5 tons of CO2, 65% of which for training the models, 18% for the international flights, 12% for data storage, and 4% for small-scale research and development experiments. To put this in perspective, the average carbon footprint per individual in the US is around 20 tons, so our project generated a little over two years of individual US emissions.

We find that the main driver of this carbon footprint is the carbon intensity of the mix used for model training. Appropriately selecting the location of calculations can significantly reduce the environmental impact of a project. For instance, in this project, running all computations in France would have reduced the total footprint to 14.9 tCO2e, 42% of which from the international flights. As the impact of the computations themselves become smaller, it is important for practitionners to more carefully weigh in exogenous contributions.

All-in-all, with careful considerations around data center choice, it is possible to run extreme-scale NLP projects with a low carbon impact.

Finally, we also identified that large-scale inference could also rapidly outtake pretraining costs in terms of carbon impact. Inference, if not centrally managed, is harder to control: with a publicly available model, it will happen on hardware decided by the end user. We thus think its equally important for practitioners to alert users regarding best efficient inference practices, and regarding best practices to limit the environmental cost of computations (e.g. choosing an efficient data center, running inference in a country with a low-impact mix, etc.)